\documentclass[letterpaper]{article} 
\usepackage[submission]{aaai23}  
\usepackage{helvet}  
\usepackage{courier}  
\usepackage[hyphens]{url}  
\usepackage{graphicx} 
\usepackage{natbib}  
\usepackage{caption} 
\usepackage{algorithm}
\usepackage{algorithmic}
\usepackage{subcaption}
\usepackage{makecell}
\usepackage{multirow}
\usepackage[normalem]{ulem}
\usepackage{newfloat}
\usepackage{listings}
\DeclareCaptionStyle{ruled}{labelfont=normalfont,labelsep=colon,strut=off} 
\floatstyle{ruled}
\newfloat{listing}{tb}{lst}{}
\floatname{listing}{Listing}
\usepackage{amsmath}
\usepackage{bbm}
\usepackage[capitalize]{cleveref}
\usepackage{color}

\newcommand\blfootnote[1]{%
  \begingroup
  \renewcommand\thefootnote{}\footnote{#1}%
  \addtocounter{footnote}{-1}%
  \endgroup
}

\title{Unifying Vision-Language Representation Space \\with Single-tower Transformer}
\author {
    Jiho Jang\textsuperscript{\rm 1}$^*$$^\dagger$ \hspace{1.6mm}
    Chaerin Kong\textsuperscript{\rm 1}$^*$$^\dagger$ \hspace{1.6mm}
    Donghyeon Jeon\textsuperscript{\rm 2} \hspace{1.6mm}
    Seonhoon Kim\textsuperscript{\rm 3}$^\dagger$ \hspace{1.6mm}
    Nojun Kwak\textsuperscript{\rm 1}\\ \vspace{1.6mm}
    $^1$Seoul National University \hspace{2mm} $^2$NAVER \hspace{2mm} $^3$Coupang\\
}

\newcommand{\nj}[1]{\textcolor{black}{#1}}
\newcommand{\jh}[1]{\textcolor{black}{#1}}

\begin{document}

\twocolumn[{
\renewcommand\twocolumn[1][]{#1}

\maketitle

\begin{center}
  \centering
  \captionsetup{type=figure}
  \vspace{-15mm}
  \includegraphics[width=0.92\textwidth]{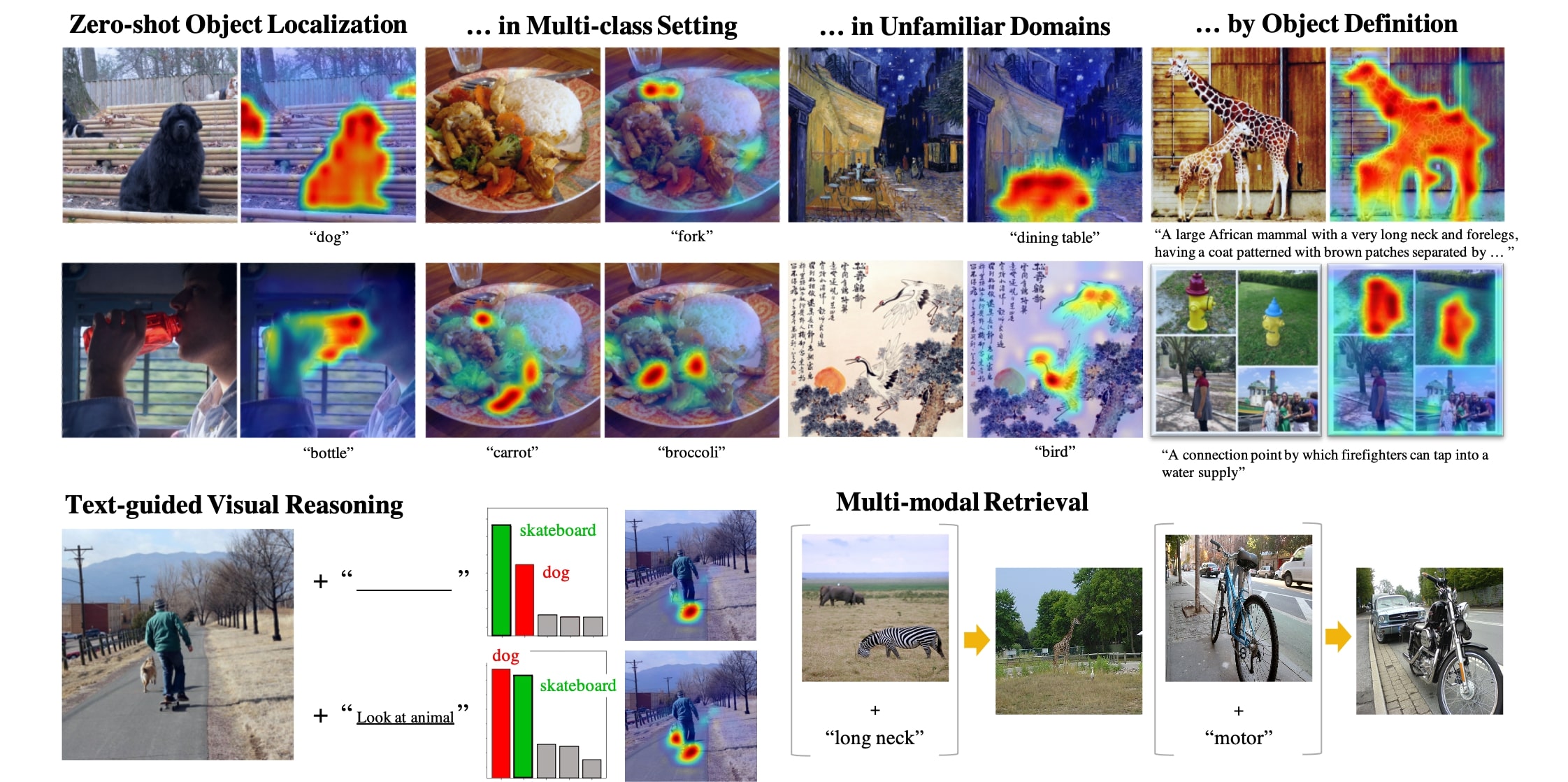}
  \captionof{figure}{
    A truly unified vision-language representation space displays intriguing properties. 
    (\textit{top}) Visualization of embedding similarities between image patches and the text prompt. (\textit{bottom left}) Steering image classification with additional text input provided as simple token sequence concatenation. Here, we plot the attention map of \texttt{[CLS]}. (\textit{bottom right}) This mixture input can also control image retrieval by combining the information from two modalities.
  }
  \label{fig:att}
\end{center}
}]

\begin{abstract}

Contrastive learning is a form of distance learning that aims to learn invariant features from two related representations. In this paper, we explore the bold hypothesis that an image and its caption can be simply regarded as two different views of the underlying mutual information, and train a model to learn a unified vision-language representation space that encodes both modalities at once in a modality-agnostic manner. We first identify difficulties in learning a generic one-tower model for vision-language pretraining (VLP), and propose OneR as a simple yet effective framework for our goal. We discover intriguing properties that distinguish OneR from the previous works that learn modality-specific representation spaces such as zero-shot object localization, text-guided visual reasoning and multi-modal retrieval, and present analyses to provide insights into this new form of multi-modal representation learning. Thorough evaluations demonstrate the potential of a unified modality-agnostic VLP framework.\blfootnote{
$^*$ These authors contributed equally. \\
$^\dagger$ Work done at NAVER.
}
\end{abstract}

\section{Introduction}

Self-supervised learning (SSL) is the core driving force behind recent boom in large scale training~\cite{devlin2018bert, radford2018improving} as it provides means to leverage a huge stack of unlabeled data handily accessible from the web. In the computer vision community, contrastive learning is one of the most popular SSL frameworks that essentially aims to maximize the mutual information between two related representations, \textit{i.e.,} views. When training with images, this is realized by first generating different views from random augmentations and encouraging the model to learn the augmentation-invariant features.

Meanwhile, the seminal work of CLIP~\cite{radford2021learning} has declared the opening of the Vision-Language Pretraining (VLP) era, where many works~\cite{li2022blip, mu2021slip, li2021align, yang2022vision, yu2022coca, yuan2021florence, zhu2022uni} have leveraged the contrastive objective for connecting images and their descriptions. However, they fundamentally differ from the aforementioned SSL contrastive framework in that they learn two separate representation spaces each for vision and language, respectively. The features from each modality are compared only after sufficient abstraction operations, typically done with \nj{self-attention layers in transformers} and separate learnable projections. This renders them short for modality-agnostic representation learning, a promising research direction towards a generic perceptual agent. 


A modality-agnostic representation learner should be capable of both 1) mapping visual and linguistic information into a unified representation space at the global sequence level and 2) mixing information within an input sequence in a modality-blind manner with generic token level attentions.
First we hypothesize that an image (\textit{e.g. a photo of panda}) and its linguistic description (\textit{e.g. the phrase ``a photo of panda"}) contains common information, which can be viewed as two different representations of implicit underlying information, analogous to the augmented views of an image. Hence, we apply contrastive SSL approach, \jh{MoCo\nj{-}v3}~\cite{chen2021empirical}, in VLP setting to congregate relevant semantics, either from visual signals, linguistic symbols, or their mixture, in a \textit{single unified representation space}. This way, our model learns to associate visual signals with structured symbols from the lowest level, breaking the boundaries between the two.

\begin{figure}[t]
\centering
  \includegraphics[width=0.44\textwidth]{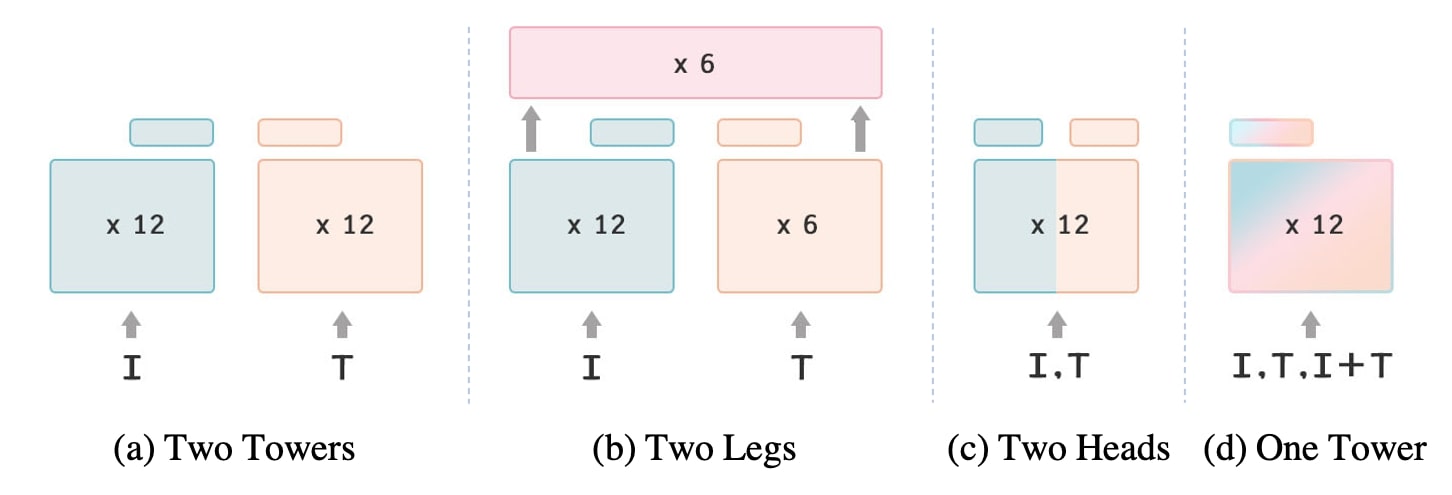}
  \caption{Typical architectures of vision-language models. (a) is the basic form, with one transformer encoder and a projector for each modality. (b) adds fusion encoder blocks on top. (c) uses a single transformer encoder, but has separate projections. (d) unifies the two modalities with a generic one-tower model (OneR).}
  \label{fig:tower}
\end{figure}

As shown in Fig.~\ref{fig:tower}, our approach is distinguished from the conventional counterparts that acknowledge the innate differences between the two modalities and encode relevant inductive bias into the model architecture. We adopt \nj{a} generic single-tower model, thus \nj{a} single representation space, to handle two different modalities at once.
We empirically demonstrate that the failure of naive single-tower image-text contrastive learning is \nj{due} to the modality gap, and propose \nj{\textit{cross-modal mixup}} as a simple yet effective remedy. Furthermore, we train our model to learn to aggregate information within each sequence in \nj{a} modality-agnostic manner by forwarding concatenation of image and text for contrastive loss computation.
This allows our model to form integrated representations even from \nj{mixed} inputs of image and text, achieving both of our previous desiderata. We name our framework OneR, short for One Representation that suits both modalities.

Aside from the academic pursuit of general intelligence, unifying multi-modal representation space with \nj{a} single generic model has been shown to have benefits in scalability and cross-modal/cross-task transferability~\cite{wang2021simvlm, mustafa2022multimodal}. We further observe that \nj{our} OneR's capacity to associate low-level visual signals to language symbols makes it an excellent zero-shot object localizer, and we can steer its visual reasoning with auxiliary language guidance thanks to its natural ability to process \textit{image}+\textit{text} mixture inputs.
The fact that mixture inputs are mapped to the same \textit{One Representation} space further renders operations like multi-modal retrieval straightforward unlike two-leg baselines (\textit{e.g.,} ALBEF~\cite{li2021align}).
We note that these properties do not rely on any modality-specific heads, segment tokens, nor special cross-attention modules, but are natural outcomes of embedding similarity and input concatenation.

Our key contributions can be summarized as:

\begin{itemize}
    \item We analyze the collapse of naive single-tower vision-language contrastive learning, and propose cross-modal mixup to mitigate the modality gap.
    \item We present OneR, a simple modality-agnostic representation learning framework that combines cross-modal mixup with contextual modality invariance to form a unified embedding space.
    \item We conduct extensive qualitative and quantitative evaluations to demonstrate the advantage of our approach, which includes distinguished capabilities in zero-shot object localization, text-guided visual reasoning and multi-modal retrieval \nj{(See Fig.~\ref{fig:att})}.
\end{itemize}

\begin{figure}[t]
\centering
  \includegraphics[width=0.38\textwidth]{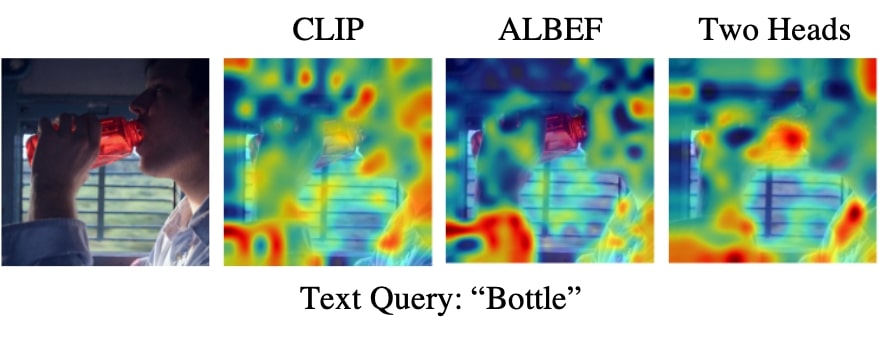}
  \caption{Patch embedding similarity map w.r.t. the text query. This clearly shows that two towers (\textit{e.g.,} CLIP), two legs (\textit{e.g.,} ALBEF) and two heads all learn modality-specific features spaces, forbidding similarity operations between embeddings. Projections are not applicable since they are only suited for the \texttt{[CLS]} token.}
  \label{fig:sim_failure}
\end{figure}

\begin{figure*}[t]
\centering
\begin{subfigure}{.3\textwidth}
  \centering
  \includegraphics[width=.9\linewidth]{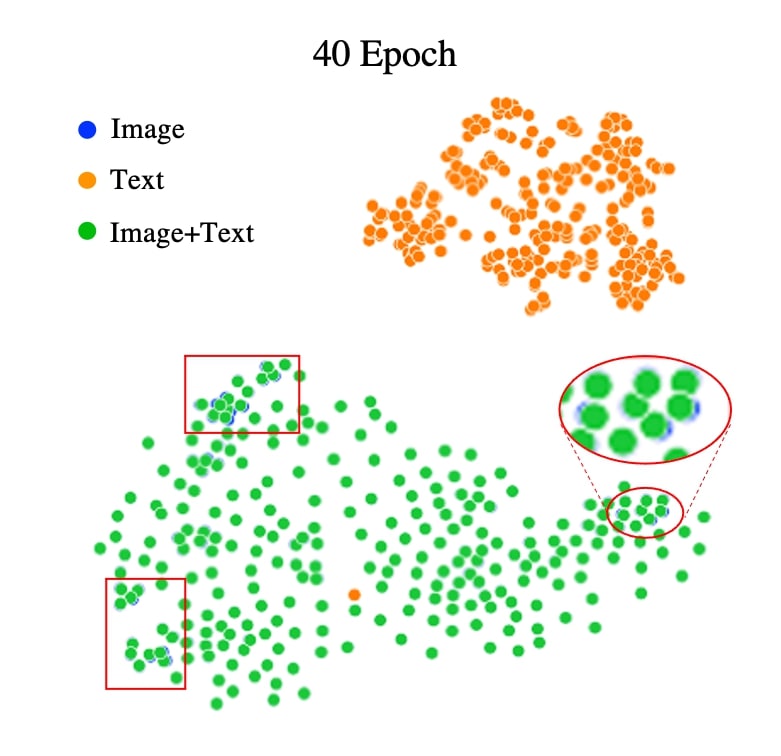}
  \caption{Naive single-tower ITC.} 
  \label{fig:tsne1}
\end{subfigure}%
\begin{subfigure}{.55\textwidth}
  \centering
  \includegraphics[width=.9\linewidth]{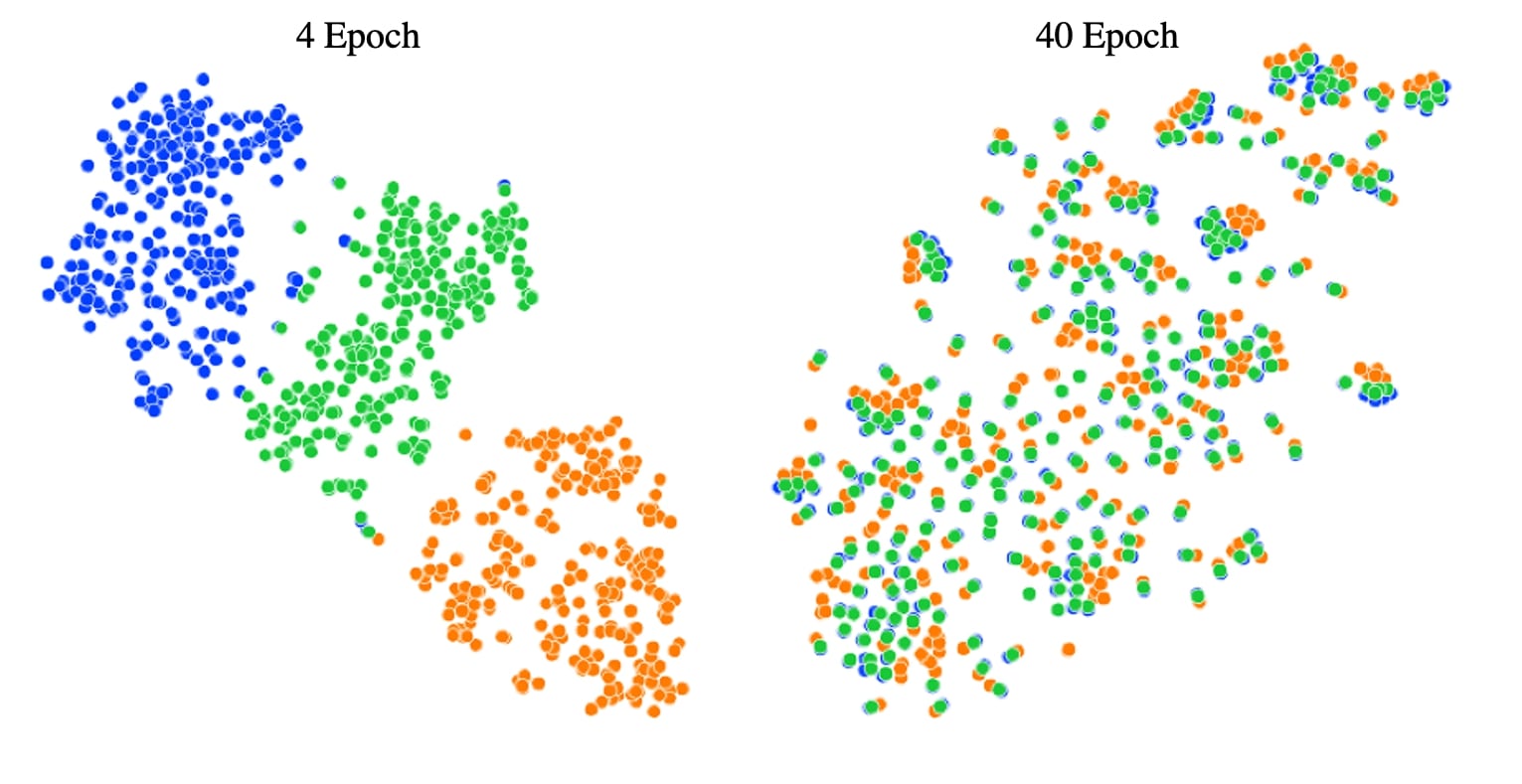}
  \caption{OneR at the beginning and the end of the training.}
  \label{fig:tsne2}
\end{subfigure}
\caption{\jh{T-SNE~\cite{van2008visualizing} representation visualization}. Single-tower model trained with naive image-text contrastive objective fails to blend two distant modalities (\textit{left}). Note that image features (blue dots) almost perfectly overlap with concatenation features (green dots), possibly due to sequence length bias \textit{(best viewed zoom-in)}. Cross-modal mixup maps embeddings from two disjoint modalities to a common middle ground, and the corresponding image, text and image+text embeddings are well clustered after 40 epochs of training (\textit{right}).
}
\label{fig:tsne}
\end{figure*}





\section{Overcoming Modality Gap with Cross-Modal Mixup}
\label{sec:2}

A typical vision-language pretraining framework with contrastive objective employs batch\nj{-}dependent InfoNCE~\cite{oord2018representation} that pulls positive \{image, text\} pairs together and pushes others apart. We state this image text contrastive \nj{(ITC)} loss as
%
%
\begin{equation}
    \mathcal{L}_{ITC} = ctr(\mathcal{F}(I), \mathcal{F}(T)),
    \label{eqn:itc}
\end{equation}

\noindent
where $ctr(A,B)=(NCE(A,B)+NCE(B,A))/2$ employs the generic InfoNCE formulation, $NCE(l, r)$, with the right term ($r$) being the EMA (exponential moving average) model output in our setting.
$\mathcal{F}(X)$ refers to the final transformer hidden state, 
and $I, T$ stands for image and text respectively. 

This formulation works well in two-tower setting\nj{s} (Fig.~\ref{fig:tower}a, \ref{fig:tower}b) with separate modality-specific encoders~\cite{radford2021learning, li2021align}, but we have observed training failure for a generic single-tower model (Fig. \ref{fig:tower}d, Tab. \ref{tab:abl_1}). Visualization of the representation space in Fig. \ref{fig:tsne1} indicates the presence of \nj{a} severe modality gap, as visual signals and linguistic symbols are significantly dissimilar. Hence, the model fails to blend these two distant modalities in a unified representation space, being unable to \nj{encode} positive \{image, text\} pair close together. 

\subsection{Cross-Modal Mixup}

Mixup~\cite{zhang2017mixup} was originally introduced in the vision community as a data-augmentation routine that improves classification performance, model robustness and generalization by extending the training data distribution with linear interpolation. Recently, a concurrent work~\cite{hao2022mixgen} has incorporated mixup into VLP in a similar spirit, applying mixup augmentation within each modality separately. \nj{Different from this, we boldly} apply mixup across modality, not as a means to augment the training data 
but as a projection to map image and text embeddings to a common middle ground. 
We find it to be an extremely simple yet effective starting point to evade the image-text modality gap, from which the traditional contrastive learning successfully guides the model for instance discrimination. The formal definition \nj{of our cross-modal mixup constrastive (XMC) loss} can be stated as
%
\begin{equation}
    \mathcal{L}_{XMC} = ctr({\mathcal{F}(I) + \mathcal{F}(T) \over 2} , {\mathcal{F}(I) + \mathcal{F}(T) \over 2}),
    \label{eqn:xmc}
\end{equation}
\noindent
where we use an online model and its momentum \nj{(EMA)} counterpart for feature extraction in practice\footnote{\nj{Note that $ctr$ by definition in \jh{\cref{eqn:xmc}} uses two separate feature extractors (online and EMA) symmetrically.}}. This straightforward approach to mitigate modality gap works surprisingly well, blending representations from the two distant modalities into a single embedding space successfully and thereby stabilizing training. Full quantitative evaluations are presented in Tab. \ref{tab:method_abl}.

\begin{table}[t]
\centering
\resizebox{0.8\columnwidth}{!}{
\begin{tabular}{l|cc}
\Xhline{4\arrayrulewidth}
Imagenet 0-shot & Top-1 Acc. & Top-5 Acc. \\ \hline
ITC             & 1.65       & 5.25       \\
ITC (two heads) & 17.46      & 35.32      \\
ITC + XMC       & 22.12      & 42.12      \\
ITC + XMC + CIC       & 22.86      & 42.88      \\
ITC + CMC       & \textbf{23.70}      & \textbf{43.15}      \\ \hline
\end{tabular}}
\caption{Zero-shot Imagenet~\cite{deng2009imagenet} evaluations. Note that all models are one tower except for the second row. Adding XMC enables one tower contrastive learning, and enforcing modality-blind token attentions further improves the performance. Masked modeling is included in all ablation models.}
\label{tab:abl_1}
\end{table}

\section{Towards Modality-Agnostic Representations}

In the previous section, we have identified modality gap as the primary obstacle for learning \nj{a} unified vision-language representation space, and proposed \nj{XMC} loss to reconcile the distant modalities. Stepping further, under the hypothesis that paired image and text contain similar information, a modality-agnostic representation should depend only on the content of the underlying information, not the modality (format; \nj{text or image}) it is expressed in.
In other words, the final embedding should be similar whether it uses image or text as the context (\textit{i.e., \nj{key and value} in self-attention}). To enforce such behavior, we devise Contextual Invariance Contrastive (CIC) loss and incorporate it into our framework.

\subsection{Contextual Modality Invariance}

The high level idea is to encourage the model representation from \nj{an} image context to be close to that from the text context. Specifically, from a pair, we choose either \nj{the} image or \nj{the} text  to be the \textit{query}. Then, at one side, we use image tokens for \textit{key} and \textit{value}, while on the other side, we use the text tokens. CIC penalizes the distance between the final representations from each side, guiding the model to extract similar information regardless of the modality of the context.
Combining it with XMC \nj{in \cref{eqn:xmc}}, the formal definition becomes
%
%
\begin{small}
\begin{equation}
    \mathcal{L}_{CIC} = ctr({\mathcal{F}(I|T) + \mathcal{F}(T|I) \over 2} , {\mathcal{F}(I|I) + \mathcal{F}(T|T) \over 2}),
    \label{eqn:cic}
\end{equation}
\end{small}
\noindent where $\mathcal{F}(X|Y)$ refers to the final embedding of $X$ \nj{(query)} given $Y$ as the context \nj{(key and value)}. We note that $\mathcal{F}(X)$ in \cref{eqn:itc} and \cref{eqn:xmc} is an abbreviated expression equivalent to $\mathcal{F}(X|X)$.


\subsection{Contextual Mixup \nj{Contrast (CMC)}}

As apparent from Tab.~\ref{tab:abl_1}, CIC improves overall performance by encouraging the model to not only embed paired image and text close together but also utilize information from image and text tokens in a similar fashion from the lowest level. To maximally leverage CIC's generic information aggregation capacity, we adapt our model for mixed modality input scenario. Formally, we replace the left contrastive term in \cref{eqn:cic} with simple concatenation of \{image, text\} \nj{($\mathcal{F}(I,T)$)} and train the model to optimize Contextual Mixup Contrastive (CMC) objective instead.
\begin{equation}
    \mathcal{L}_{CMC} = ctr(\mathcal{F}(I,T|I,T), {\mathcal{F}(I|I) + \mathcal{F}(T|T) \over 2})
    \label{eqn:cmc}
\end{equation}
This is a generalized form that further integrates XMC and CIC, which explicitly 
guides the model to embed mixed modality inputs to the unified V-L representation space after adequate integration of information from two different modalities.
We utilize this property for text-aided visual reasoning (\cref{tab:boot}) and multi-modal retrieval (\cref{fig:att}).
The high-level idea is that the self-attention feature of concatenated input can be roughly decomposed to self-attention feature of each plus the cross-attention features, and the theoretical verification is provided in the supplementary.

\subsection{\nj{One Representation (OneR)}}

\begin{figure}[t]
\centering
  \includegraphics[width=0.44\textwidth]{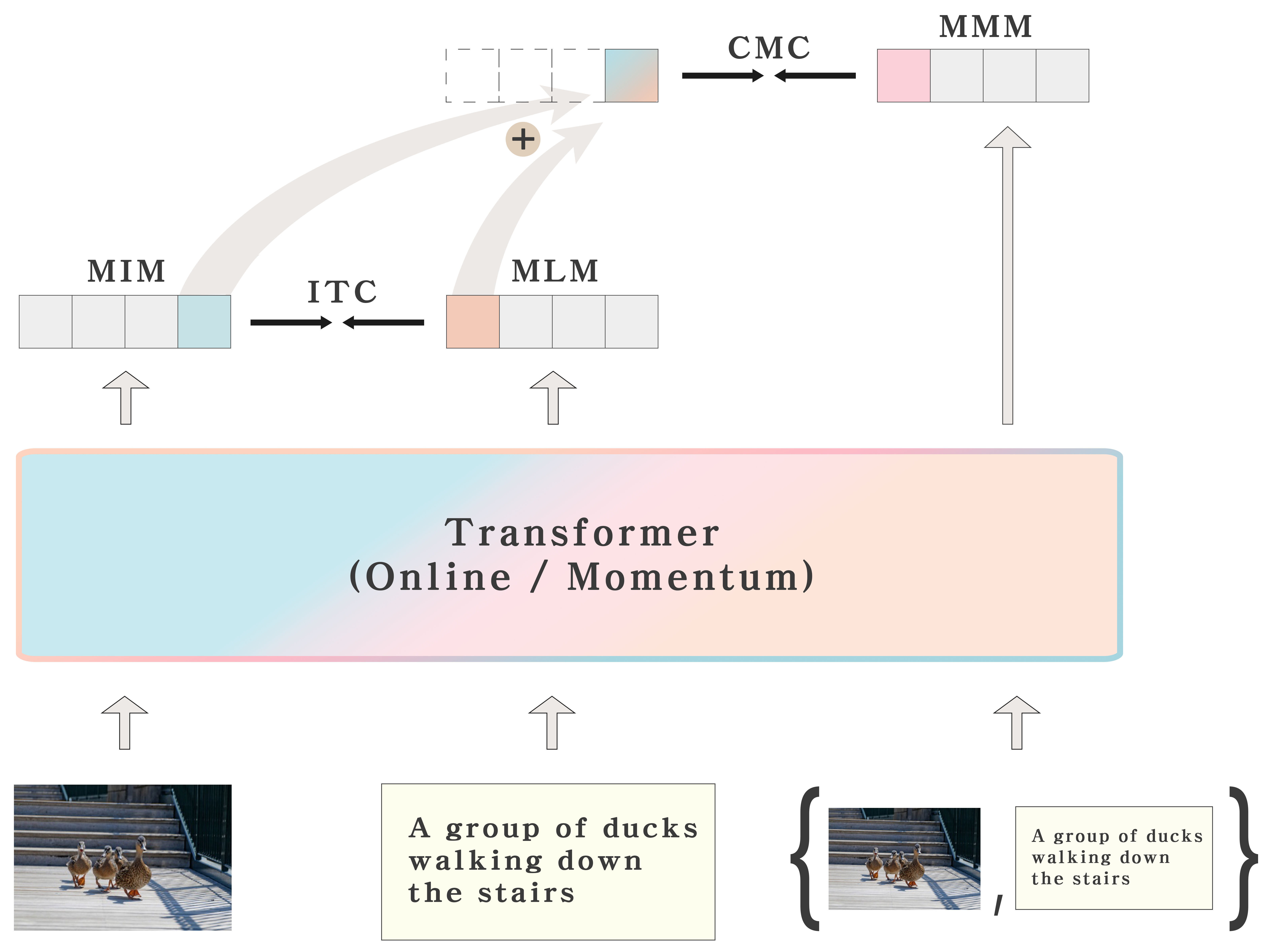}
  \caption{Overview of OneR. Image-text contrastive and contextual mixup contrastive objective provide guidance in parallel with masked modeling for three input types: image, language and multi-modal (image+text).}
  \label{fig:overview}
\end{figure}

\begin{table}[t]
\centering
\resizebox{0.85\columnwidth}{!}{
\begin{tabular}{l|cc}
\Xhline{4\arrayrulewidth}
\multicolumn{1}{c|}{Method} & \multicolumn{2}{c}{Formulation} \\ \hline
ITC    & $\mathcal{F}(I)$ & $\mathcal{F}(T)$ \\
XMC    & $(\mathcal{F}(I)+\mathcal{F}(T))/2$ & $(\mathcal{F}(I)+\mathcal{F}(T))/2$       \\
CIC    & $(\mathcal{F}(I|T)+\mathcal{F}(T|I))/2$ & $(\mathcal{F}(I)+\mathcal{F}(T))/2$      \\
CMC    & $\mathcal{F}(I,T|I,T)$ & $(\mathcal{F}(I)+\mathcal{F}(T))/2$     \\ \hline
\end{tabular}}
\caption{Summary of the contrastive objectives.}
\label{tab:ctr}
\end{table}

\nj{Fig.~\ref{fig:overview}} illustrates the overall pipeline of OneR. Model input can be one of \textit{image}, \textit{text} or \textit{image}+\textit{text}, and \nj{CMC} objective \nj{in \cref{eqn:cmc}} is combined with the traditional image-text contrastive (ITC) loss. Masked modeling is also carried out for all three input types (\textit{i.e., image, text and multi-modal}). Our framework employs no modality-specific architectural component except for the initial token embedding layer, making our model generic and modality-agnostic with minimal inductive bias. Tab.~\ref{tab:ctr} summarizes the overall formulations.

\section{Experiments}

\subsection{Training Setup}
\noindent
\textbf{Datasets} Following prior works~\cite{li2021align, yang2022vision, gan2020large}, we train OneR on the combination of CC3M~\cite{sharma2018conceptual}, SBU Captions~\cite{ordonez2011im2text}, Visual Genome~\cite{krishna2017visual} and COCO~\cite{lin2014microsoft}, which sums up to 4M images and 5.1M image-text pairs. Ablation models are trained on CC3M.

\begin{figure*}[t]
\centering
  \includegraphics[width=0.75\textwidth]{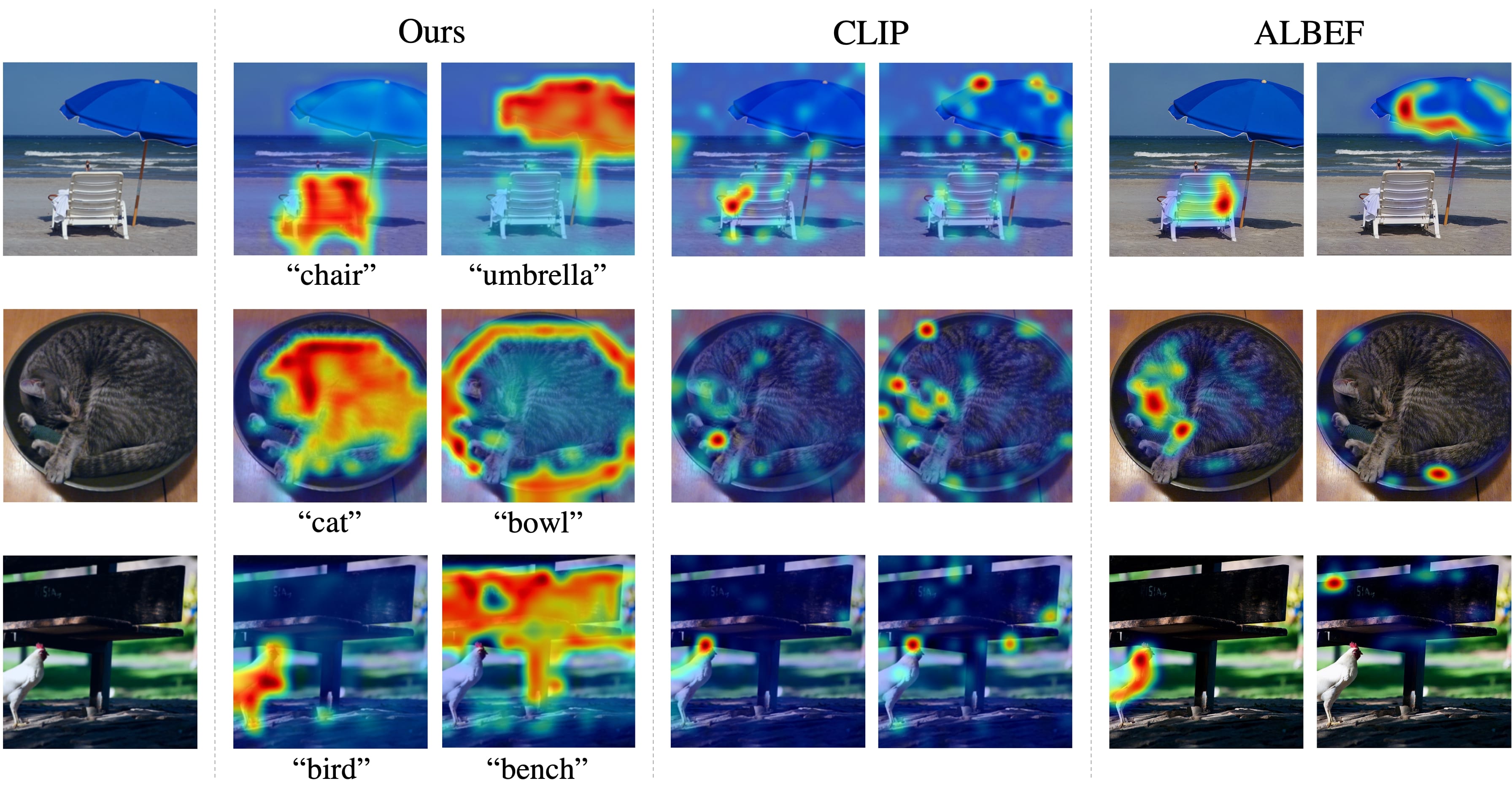}
  \caption{Qualitative evaluation for object-level scene understanding. We simply compute token similarities for OneR, and Grad-CAM is used for CLIP and ALBEF. It is visually apparent that OneR correctly associates low-level visual signals to its corresponding language symbol, resulting in segmentation-map-like patch similarity maps.}
  \label{fig:qual}
\end{figure*}

\noindent
\textbf{Implementation Details} 
We adopt the model architecture of Masked AutoEncoder~\cite{he2022masked} with BERT~\cite{devlin2018bert} word embeddings and language modeling head. Unlike most prior works on VLP, we initialize our entire model \textit{from scratch}, as neither ViT nor \nj{BERT} suits our goal towards a unified VL representation space~\footnote{Two-legged models typically initialize their encoders with a pretrained ViT and a \nj{pretrained} language model such as \nj{BERT}, which makes the \nj{training} much simpler.}. \nj{1D and 2D} sinusoidal positional embeddings are added to text and image respectively, and a single \texttt{[CLS]} token is prepended to all three input types. 
Special modality indicator tokens (\textit{e.g.,} \texttt{[SEP]} or \texttt{[SEG]}) are further removed from typical one tower baselines in order to train a fully modality-agnostic representation learner. We train our model with 32 A100 GPUs for 40 epochs under PyTorch framework. Details on hyperparameters are listed in the supplementary.


\subsection{Properties of One Representation}

\noindent
\textbf{Zero-shot Localization}
Conventional vision-language transformers typically rely on \texttt{[CLS]} cross-attention map or Grad-CAM~\cite{selvaraju2017grad} for visualization. However, the former attributes the global semantics to each local region, rendering it unsuitable for complex scene understanding such as multi-class localization (\cref{fig:att}), while the latter requires a separately devised procedure that involves back propagation. 
One of the most distinguished qualities of OneR is its natural proficiency for object localization. Throughout the paper, we simply compute the \textit{cosine similarities} between image patch embeddings and the average-pooled text embedding for visualization. This is possible only because OneR maps both visual and textual information to a unified embedding space where their feature similarity correctly indicates the semantic relevance. Otherwise, the token level similarity map conveys no meaningful information, as illustrated in \cref{fig:sim_failure}.

We present qualitative comparison on zero-shot localization with two competitive baselines, CLIP and ALBEF, where Grad-CAM is used for their visualizations as it yields the best output. Looking at \cref{fig:qual}, we can see that Grad-CAM of ALBEF better captures the spatial details compared to CLIP, but OneR has the most fine-grained visual reasoning, resulting in almost segmentation\nj{-}map-like patch similarity maps. This clearly shows that OneR has the capacity to relate low-level visual signals to their corresponding linguistic concepts in a unified vision-language representation space.

\begin{table}[t]
\resizebox{\columnwidth}{!}{
\begin{tabular}{l|cc|cc}
\hline
\Xhline{4\arrayrulewidth}
\multirow{2}{*}{Bootstrapped Language Guidance} & \multicolumn{2}{c|}{ImageNet 0-shot} & \multicolumn{2}{c}{CIFAR100 0-shot} \\
                                                & top-1            & top-5            & top-1            & top-5            \\ \hline
OneR (4M)                                       & 27.33            & 50.17            & 31.45            & 57.52            \\
OneR-Bootstrapped (4M)                          & \textbf{28.00}   & \textbf{50.69}   & \textbf{32.23}   & \textbf{58.24}   \\ \hline
\end{tabular}}
\caption{Evaluation with bootstrapped language guidance. We can feed predicted class labels in simple concatenation to the input image to further improve accuracy. Note that this is not possible with two-tower or two-leg models, as the former does not accept mixture inputs and the latter forms a separate feature space after fusion, forbidding the similarity operation.}
\label{tab:boot}
\end{table}

\begin{table}[t]
\resizebox{\columnwidth}{!}{
\begin{tabular}{l|c|c|ccc}
\hline
\Xhline{4\arrayrulewidth}
\multirow{2}{*}{Cross-modal Transfer}      & \multirow{2}{*}{Architecture} & 0-shot INet & \multicolumn{2}{c}{MS COCO}                                      \\
                                           &                            & top-1 & TR@1            & IR@1          \\ \hline
\multirow{2}{*}{SBU}                       & two heads                  & \textbf{7.28} & \textbf{8.88}  & 5.73          \\
                                           & one tower                  & 6.49 & 8.60           & \textbf{5.77} \\ \hline
\multirow{2}{*}{SBU + CC3M (caption only)} & two heads                  & \textbf{8.59} & 10.41          & 6.87          \\
                                           & one tower                  & 8.54 & \textbf{11.31} & \textbf{7.20} \\ \hline
\multirow{2}{*}{Gain}                      & two heads                  & 1.31 & 1.53           & 1.14          \\
                                           & one tower                  & \textbf{2.07} & \textbf{2.71}  & \textbf{1.43} \\ \hline
\end{tabular}
}
\caption{Cross-modal knowledge transfer. Under a unified representation space, additional training in one modality benefits performance in the other modality with bigger margins.}
\label{tab:xmodal}
\end{table}

\begin{table*}[t]
\centering
\resizebox{0.98\linewidth}{!}{
\begin{tabular}{l|ccc|cccccc|cccccc}
\Xhline{4\arrayrulewidth}
\multicolumn{1}{c|}{\multirow{3}{*}{Method}} & \multirow{3}{*}{Architecture} & \multirow{3}{*}{Pre.} & \multirow{3}{*}{\#Images} & \multicolumn{6}{c|}{Zero-shot MS-COCO (5K)}                                         & \multicolumn{6}{c}{Fine-tuned MS-COCO (5K)}                                       \\
\multicolumn{1}{c|}{}  &                               &                           &                           & \multicolumn{3}{c}{Text Retrieval} & \multicolumn{3}{c|}{Image Retrieval} & \multicolumn{3}{c}{Text Retrieval} & \multicolumn{3}{c}{Image Retrieval} \\
\multicolumn{1}{c|}{}  &                               &                           &                           & R@1        & R@5       & R@10      & R@1        & R@5        & R@10       & R@1        & R@5       & R@10      & R@1        & R@5        & R@10      \\ \hline
ImageBert$^\dagger$                                    & One Tower     & O             & 6M                        & 44.0       & 71.2      & 80.4      & 32.3       & 59.0       & 70.2       & \textbf{66.4}       & \textbf{89.8}      & \textbf{94.4}      & \textbf{50.5}       & \textbf{78.7}       & \textbf{87.1}      \\
ViLT                                         & One Tower     & O             & 4M                        & 56.5       & 82.6      & 89.6      & 40.4       & 70.0       & 81.1       & 61.5       & 86.3      & 92.7      & 42.7       & 72.9       & 83.1      \\
Uni-Perceiver                                & One Tower     & X             & 44.3M                     & \underline{57.7}       & \underline{85.6}      & \underline{92.3}      & \underline{46.3}       & \textbf{75.0}       & \underline{84.0}       & 64.7       & \underline{87.8}      & \underline{93.7}      & \underline{48.3}       & 75.9       & 84.5      \\
OneR                                         & One tower     & X             & 4M                        & \textbf{62.9}       & \textbf{86.3}      & \textbf{92.5}      & \textbf{47.0}       & \underline{74.7}       & \textbf{84.1}       & \underline{66.1}       & \underline{87.8}      & 93.2      & \underline{48.3}       & \underline{76.0}       & \underline{85.2}      \\ \hline
CLIP                                         & Two towers    & X             & 400M                      & 58.4       & 81.5      & 88.1      & 37.8       & 62.4       & 72.2       & -          & -         & -         & -          & -          & -         \\
FLAVA                                        & Two legs      & O             & 70M                       & 42.7       & 76.8      & -         & 38.4       & 67.5       & -          & -          & -         & -         & -          & -          & -         \\
ALBEF                                        & Two legs      & O             & 4M                        & 68.7       & 89.5      & 94.7      & 50.1       & 76.4       & 84.5       & 73.1       & 91.4      & 96.0      & 56.8       & 81.5       & 89.2      \\
TCL                                          & Two legs      & O             & 4M                        & 71.4       & 90.8      & 95.4      & 53.5       & 79.0       & 87.1       & 75.6       & 92.8      & 96.7      & 59.0       & 83.2       & 89.9      \\ \hline
\end{tabular}}
\caption{Quantitative evaluations on COCO image and text retrieval. Two-legs models generally perform better as they have modality-specific encoders and more parameters. Note that previous vision-language models typically initialize their weights from a pretrained model such as Imagenet ViT or Bert to help training (\textit{Pre.}). OneR, on the other hand, achieves the best zero-shot performance among one-tower models without any initialization prior, and compares on par after fine-tuning. $\dagger$ indicates the use of an additional object detection module.}
\label{tab:quant}
\end{table*}



\noindent
\textbf{Text-guided Visual Reasoning}
As illustrated in \cref{fig:att}, OneR's ability to understand \textit{image+text} mixture input opens up possibilities for diverse forms of multi-modal reasoning. For example, we can simply concatenate additional text to the image input sequence to guide its visual representation, which can be particularly useful in a complex scene understanding setting where an image contains more than one dominant semantic. In such cases, we can \textit{tell} the model where to focus to suit our goals. We provide quantitative results to further demonstrate this property in \cref{tab:boot}, where we bootstrap with language guidance to improve zero-shot classification accuracy. Specifically, for each image, we retrieve top-10 class labels upon embedding similarity. Then we concatenate each to the image sequence and compute similarity once more, similar to sample re-ranking. The intuition is that when \textit{image+text} input is given, image patches that attend strongly to the text label are strengthened by the attention mechanism, resulting in clearer representations. We note that we do not provide any external guidance during this procedure, which makes these gains essentially \textit{free}.

\noindent
\textbf{Cross-modal Knowledge Transfer}
We hypothesize that under a unified vision-language representation space, additional training on one modality should benefit performance in the other modality. \cref{tab:xmodal} validates our conjectures, as additional training with language data results in greater gains for the unified one-tower model. This could indicate better scalability of one-tower models, as there is much more single-modality data available than image-text pairs in the web, which we leave for future works.

\subsection{Quantitative Evaluations}

\cref{tab:quant} shows the quantitative comparison with state-of-the-art methods on widely used image-text retrieval benchmark. Models with modality-specific encoders typically show better performance as they have more parameters and architectural inductive bias. Among one-tower baselines, OneR shows the best zero-shot performance, sometimes with significant margins. 
We note that OneR achieves such competent outcome without any initialization prior commonly used in the literature.
This shows that vision and language modalities \textit{can} be effectively encoded in a single representation space with minimal inductive bias, once the aforementioned obstacle (\textit{i.e.,} innate modality gap) is overcome.

In \cref{tab:method_abl}, we present full ablations for our framework. Naive ITC with one tower fails due to the modality gap, and adding modality-specific projectors can be the minimal architectural modification that works, but still lags behind our method. CMC combines XMC and CIC into a concise formulation, which is explained further in the supplementary, resulting in the best performance that surpasses competent two-tower baselines.

\begin{table}[t]
\centering
\resizebox{0.95\columnwidth}{!}{
\begin{tabular}{l|ccccc}
\Xhline{4\arrayrulewidth}
\multirow{2}{*}{Method} & Imagenet      & \multicolumn{4}{c}{MS-COCO}                                                                   \\
                        & Top-1 Acc.    & TR@1          & TR@5          & IR@1          & IR@5          \\ \hline
CLIP                    & 17.1          & 15.0          & 34.8          & 10.9          & 26.7          \\
SLIP                    & 23.0          & 21.7          & 45.1          & 15.6          & 35.2          \\ \hline
ITC                     & 1.6           & 0.8           & 2.5           & 0.7           & 2.2           \\
ITC (two heads)         & 17.5          & 10.4          & 26.8          & 10.7          & 26.4          \\
ITC + XMC               & 22.1          & 25.2          & 48.1          & 15.2          & 33.6          \\
ITC + XMC + CIC               & 22.9          & 25.4          & 48.1          & 16.3          & 35.5         \\
ITC + CMC (OneR)        & \textbf{23.7} & \textbf{25.5} & \textbf{48.2} & \textbf{16.9} & \textbf{36.9} \\ \hline
\end{tabular}}
\caption{Method ablation. Our proposed components consistently improve the performance, with the final CMC outperforming the two-tower baseline that uses more parameters and intra-modal contrastive loss. Additional ablations are presented in the supplementary.}
\label{tab:method_abl}
\end{table}

\subsection{Visual Reasoning Analysis}

We further analyze the visual reasoning mechanism of OneR to provide insights into the properties of unified vision-language representation space.

\begin{figure}[t]
\centering
  \includegraphics[width=0.35\textwidth]{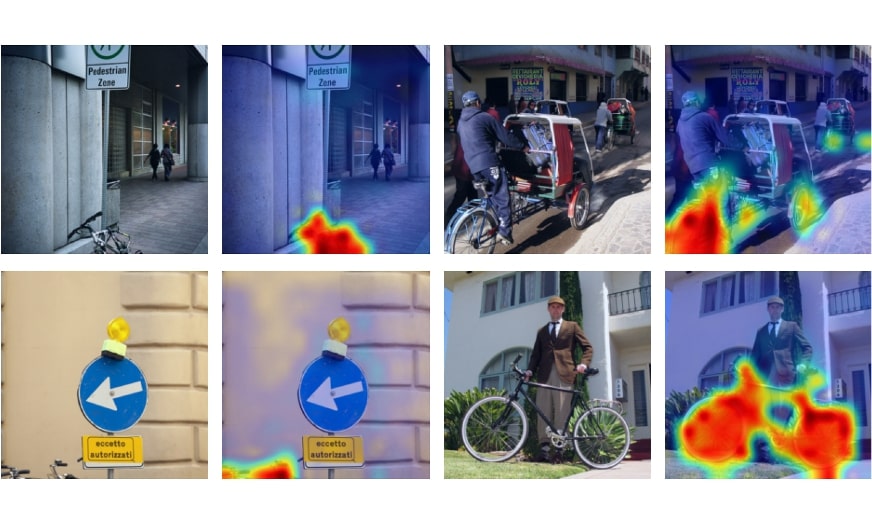}
  \caption{As OneR learns to associate low-level visual signals to the language, \nj{it} shows robust visual reasoning even with \nj{a} relatively small pretraining dataset. Above, OneR robustly recognizes \nj{\textit{bicycle}} from different visual clues (\textit{e.g., handles, wheels or the body}).}
  \label{fig:robustness}
\end{figure}

\noindent
\textbf{Robustness}
\cref{fig:robustness} shows an example of how OneR recognizes an object (bicycle, in this case) with different visual clues. OneR recognizes a bicycle even from partial images of handles or wheels, which we believe is key to its robustness in visual understanding. We present additional results in the supplementary materials, including inference in unfamiliar domains. 

\begin{figure}[t]
\centering
  \includegraphics[width=0.49\textwidth]{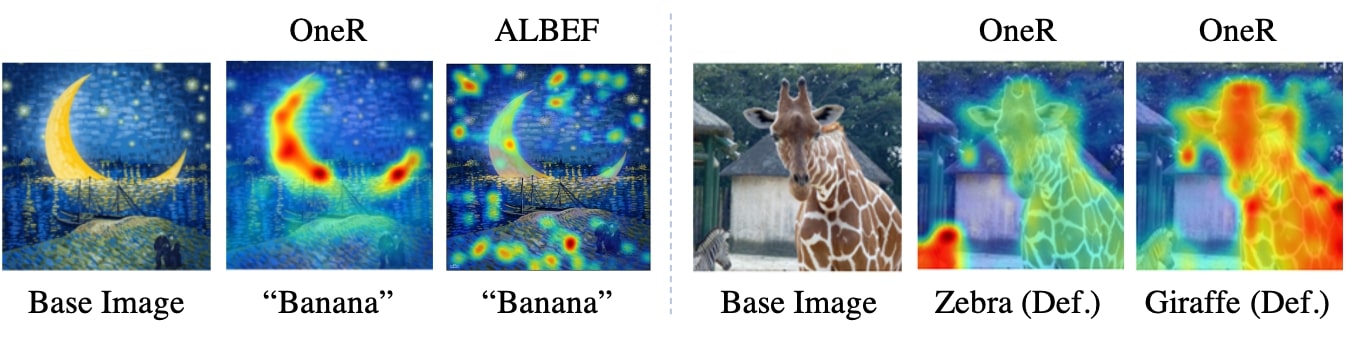}
  \caption{(\textit{left}) Patch embedding similarity \nj{(OneR)} and Grad-Cam \nj{(ALBEF)}. (\textit{right}) Patch embedding similarity map \nj{w.r.t.} definitions of zebra and giraffe.
  }
  \label{fig:multilevel}
\end{figure}

\noindent
\textbf{Multi-level vision-language connection}
Looking at Fig.~\ref{fig:multilevel}, OneR recognizes the \textit{moon} as being visually similar to \textit{banana} in terms of embedding similarity, while ALBEF condenses the global semantic in \texttt{[CLS]}, resulting in a randomly scattered Grad-CAM. Although this can be viewed as a failure case of OneR, it reveals how OneR perceives the visual signals. \nj{On} the right, we can see that \textit{zebra} and \textit{giraffe} are visually similar, and their definitions contain similar phrases such as `an African mammal', resulting in some overlaps in the two similarity maps. However, after abstracting the linguistic semantics, the model correctly identifies each, which shows its ability to process high-level semantics as well. Overall, OneR learns both low-level and high-level vision-language connections, making it a competent modality-agnostic representation learner.

\section{Related Works}

\noindent
\textbf{Vision-Language Pretraining}
Motivated by the success in self-supervised learning, Vision-Language Pretraining has become a popular research topic. CLIP~\cite{radford2021learning} first demonstrate\nj{d} that contrastive learning framework equipped with large scale paired image-text dataset can reach the performance comparable to \nj{those of} fully supervised baselines. ALIGN~\cite{jia2021scaling} scale\nj{d} up the training with noisy image\nj{s} and alt-text pair \nj{data}. Another line of works~\cite{li2020oscar, li2020unimo, chen2019uniter, gan2020large} leverage\nj{d} an off-the-shelf object detector to extract visual concepts first, which \nj{were then} used to train the multi-modal transformer. 
In an attempt to learn cross-modal interactions, ALBEF~\cite{li2021align}, TCL~\cite{yang2022vision}, FLAVA~\cite{singh2022flava}, and Florence~\cite{yuan2021florence} adopt\nj{ed} multi-modal fusion layers on top of modality-specific transformer encoders. These models show impressive performance on various vision-language tasks, as they are capable of processing both single-modal and multi-modal inputs. Another group of works~\cite{li2022blip, yu2022coca, wang2021simvlm, mokady2021clipcap} explore\nj{d} generative modeling, typically in the form of image captioning, to further improve performances on challenging tasks such as visual question answering. 


\noindent
\textbf{Unified VL Framework}
Along with the efforts to push the state-of-the-art further, some works \nj{have focused} on training a generic model that can handle diverse problem scenarios with minimal human inductive bias. Uni-Perceiver~\cite{zhu2022uni} adopt\nj{ed} a single-tower transformer architecture to tackle different V-L tasks. Unified-IO further unifie\nj{d} the input/output format using a pretrained VQ-VAE, modeling a wide range of tasks with a sequence-to-sequence framework. These works have demonstrated promising direction towards a unified perception system, but they all employ the multi-task pretraining strategy, which pools different task datasets together to train the network. This approach can be less scalable compared to simpler contrastive frameworks that only leverage weakly linked image-text pairs such as CLIP and ALIGN. UFO~\cite{wang2021ufo} has shown that a single transformer model suffice\nj{s} for typical vision-language pretraining, but falls short towards a \textit{unified vision-language representation space} as they attach two independent projectors to map the modalities together. LIMoE~\cite{mustafa2022multimodal}, a concurrent work of ours, also explores single-tower (two heads) VLP but with a new set of inductive biases, \textit{i.e.,} mixture of experts, encoded into the architecture. OneR, in contrast, learns a common embedding space without any modality-specific projections, which empowers the model with unique capabilities previously demonstrated.


\noindent
\textbf{Self-supervised Learning}
Self-supervised learning first bloomed in the NLP domain as masked language modeling (MLM) and language modeling (LM) enabled \nj{pretraining large language models} with huge stock of unlabeled text corpus~\cite{devlin2018bert, radford2018improving, lewis2019bart, liu2019roberta}.
In the vision community, contrastive learning has led the rise of SSL. MoCo~\cite{he2020momentum} and SimCLR~\cite{chen2020simple} are the pioneers to demonstrate the potential of contrastive representation learning, which we adapt for VLP setting. BYOL~\cite{grill2020bootstrap} and SimSiam~\cite{chen2021exploring} explore\nj{d} new setting\nj{s} with no negative samples that \nj{mitigate} the batch size dependency. Recent works~\cite{caron2021emerging, chen2021empirical, jang2021self} actively employ ViT~\cite{dosovitskiy2020image} to improve the performance and discover new properties. This architecture is widely used in VLP as it can model data from different modalities in an elegant and integrated manner.



\section{Conclusion}

Modality-agnostic representation learning is a meaningful step towards a generic perceptual agent that understands the environment in a similar way as humans do. In this work, we explore the difficulties of unifying modalities into a single representation space, and introduce OneR as a generic framework that shows unique qualities as a modality-agnostic representation learner. 

\noindent
\textbf{Acknowledgements}
This work was supported by NRF grant (2021R1A2C3006659) and IITP grants (No.2022-0-00953, No.2021-0-01343), all of which are funded by Korean Government.

\bibliography{aaai23.bib} 

%
\lstset{%
	basicstyle={\footnotesize\ttfamily},
	numbers=left,numberstyle=\footnotesize,xleftmargin=2em,
	aboveskip=0pt,belowskip=0pt,%
	showstringspaces=false,tabsize=2,breaklines=true}
\floatstyle{ruled}
\newfloat{listing}{tb}{lst}{}
\floatname{listing}{Listing}
\newfloat{footnote}
%
\pdfinfo{
/TemplateVersion (2023.1)
}

\setcounter{secnumdepth}{0} 

\title{Unifying Vision-Language Representation Space \\with Single-tower Transformer\\-- Supplementary Materials --}

\newcommand{\blue}[1]{\textcolor{blue}{#1}}
\newcommand{\red}[1]{\textcolor{red}{#1}}

\renewcommand\thesection{\Alph{section}}
\newcommand{\beginsupplement}{%
        \setcounter{table}{0}
        \renewcommand{\thetable}{S\arabic{table}}%
        \setcounter{figure}{0}
        \renewcommand{\thefigure}{S\arabic{figure}}%
     }


\twocolumn[{%
\renewcommand\twocolumn[1][]{#1}%
\section{A. Theoretical Explanation of CMC}
\:\\
\begin{center}
  \centering
  \captionsetup{type=figure}
  \includegraphics[width=0.96\textwidth]{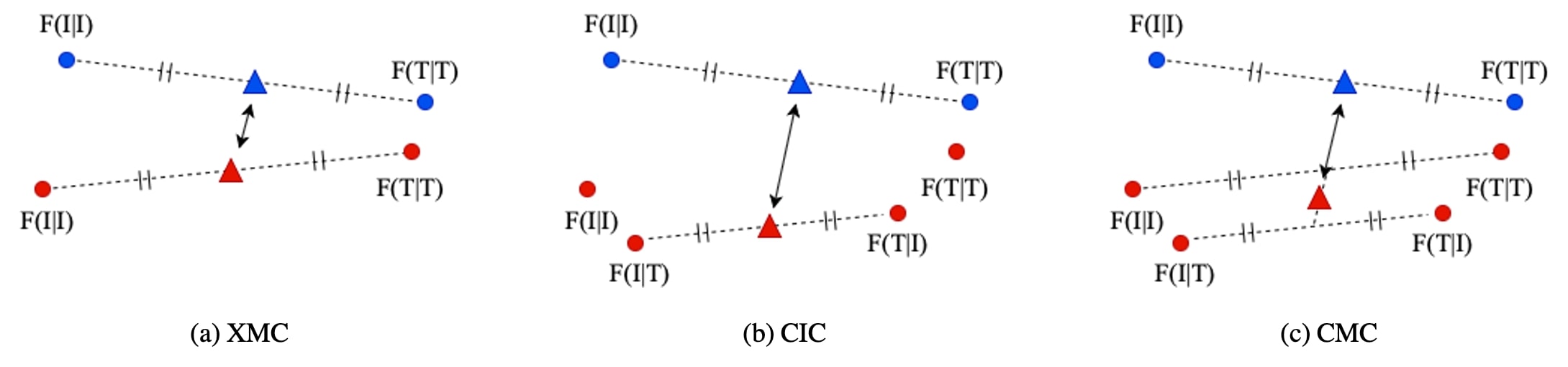}
  \captionof{figure}{
    Graphical illustration of the proposed contrastive components. \blue{Blue} dots represent the momentum features and \red{red} dots indicate the online network features. Note that these can be swapped in practice.
  }
  \label{fig:ctrs}
\end{center}

\:\\
In this paper, we propose XMC and CIC loss as

\begin{equation}
    \mathcal{L}_{XMC} = ctr({\mathcal{F}(I|I) + \mathcal{F}(T|T) \over 2}, {\mathcal{F}(I|I) + \mathcal{F}(T|T) \over 2})
\end{equation}

\begin{equation}
    \mathcal{L}_{CIC} = ctr({\mathcal{F}(I|T) + \mathcal{F}(T|I) \over 2}, {\mathcal{F}(I|I) + \mathcal{F}(T|T) \over 2}).
\end{equation}
We combine these two components to obtain the concise formulation of CMC as illustrated below.

\begin{equation}
    \frac{\mathcal{L}_{XMC}+\mathcal{L}_{CIC}}{2} \simeq
    \mathcal{A} ( \mathcal{L}_{XMC} , \mathcal{L}_{CIC} ) =  ctr({\mathcal{F}(I|I) + \mathcal{F}(I|T) + \mathcal{F}(T|T) + \mathcal{F}(T|I) \over 4}, {\mathcal{F}(I|I) + \mathcal{F}(T|T) \over 2})
\end{equation}

where $\mathcal{A}(\cdot,\cdot)$ is a kind of average operation (this will be different from arithmetic mean, harmonic mean or geometric mean) which will be approximately equal to the arithmetic mean.

Then, we define the attention module $f(X|Y)$ as

\begin{equation}
    f(X|Y) = S({Q_X K_Y^T \over \sqrt{d_k}})V_Y,
\end{equation}

where S is the softmax operation along each row.

\begin{align}
    f(X|X,Y)&=S({Q_X cat(K_X,K_Y)^T \over \sqrt{d_k}})cat(V_X,V_Y)\\
            &=S({cat(Q_X K_X^T,Q_X K_Y^T) \over \sqrt{d_k}})cat(V_X,V_Y)\\
            &=cat(\lambda_X S({Q_X K_X^T \over \sqrt{d_k}}), (\nj{I}-\lambda_X) S({Q_X K_Y^T \over \sqrt{d_k}}))cat(V_X,V_Y)\\
            &=\lambda_X S({Q_X K_X^T \over \sqrt{d_k}})V_X + (\nj{I}-\lambda_X) S({Q_X K_Y^T \over \sqrt{d_k}})V_Y\\
            &=\lambda_X f(X|X) + (\nj{I}-\lambda_X) f(X|Y)
\end{align}

}]

\onecolumn
Here, $cat(\cdot,\cdot)$ is the concatenate operation, \nj{$I$ is the identity matrix,} and $\lambda_X$ is a diagonal matrix which can be defined as

\begin{equation}
    \lambda_X^{(i)} = {\sum_{j=1}^{l_X} exp(Q_X^{(i)}K_X^{(j)}) \over \sum_{j=1}^{l_X} exp(Q_X^{(i)}K_X^{(j)}) + \sum_{j=1}^{l_Y} exp(Q_X^{(i)}K_Y^{(j)}) }
\end{equation}

\begin{equation}
    \lambda_X = \begin{bmatrix}
    \lambda_X^{(1)} & & \\
    & \ddots & \\
    & & \lambda_X^{(l_X)}
  \end{bmatrix}
\end{equation}

\noindent
with $l_X$ and $l_Y$ being the sequence length of X and Y, respectively. So far, we have decomposed the softmax of concatenated input into weighted sum of two terms.
If we consider the final transformer output $\mathcal{F}(X|Y)$ as the average pooling of the attention module $f(X|Y)$,

\begin{equation}
    \mathcal{F}(X|Y) = {1 \over l_X}\text{\Large{$\mathbbm{1}$}}^T f(X|Y),
\end{equation}

\noindent
\nj{where $\mathbbm{1}$ is the all-one vector}, then we can further decompose the final self-attention output of the concatenated input as four different terms with corresponding weights as follows:

\begin{align}
    \mathcal{F}(X,Y|X,Y)
    &= {1 \over l_X + l_Y}\text{\Large{$\mathbbm{1}$}}^T f(X,Y|X,Y) \\
    &= \alpha{1 \over l_X}\text{\Large{$\mathbbm{1}$}}^T f(X|X,Y) + (1-\alpha){1 \over l_Y}\text{\Large{$\mathbbm{1}$}}^T f(Y|X,Y)\\
    &=\alpha\lambda_X{1 \over l_X}\text{\Large{$\mathbbm{1}$}}^T f(X|X) + \alpha(1-\lambda_X){1 \over l_X}\text{\Large{$\mathbbm{1}$}}^T f(X|Y) \\
    &\;\;\;+(1-\alpha)\lambda_Y{1 \over l_Y}\text{\Large{$\mathbbm{1}$}}^T f(Y|Y) + (1-\alpha)(1-\lambda_Y){1 \over l_Y}\text{\Large{$\mathbbm{1}$}}^T f(Y|X) \\
    &=\beta_1\mathcal{F}(X|X) + \beta_2\mathcal{F}(X|Y) + \beta_3\mathcal{F}(Y|Y) + \beta_4\mathcal{F}(Y|X).
\end{align}

\noindent
Here, $\alpha$ is the sequence length ratio ${l_X \over l_X + l_Y}$, and $\beta_1 + \beta_2 + \beta_3 + \beta_4 = 1$. Note that we substitute Eq.(9) into Eq.(14) to obtain the result.
In order to simplify the formulation, we assume $\beta_1 = \beta_2 = \beta_3 = \beta_4$ in practice, which allows substituting Eq.(17) to Eq.(3) to obtain the final equivalence.

\begin{equation}
    \mathcal{A} ( \mathcal{L}_{XMC} , \mathcal{L}_{CIC} ) \simeq ctr(\mathcal{F}(I,T|I,T), {\mathcal{F}(I|I) + \mathcal{F}(T|T) \over 2}) = \mathcal{L}_{CMC}.
\end{equation}

We note that this is not a rigorous theoretical proof for our CMC formulation. Rather, it is to show how we combine XMC with CIC to formulate CMC, a concise form that explicitly trains the model for mixed-modality input scenario. 


\section{B. Additional Ablations}

\begin{table}[h]
\centering
\resizebox{0.8\columnwidth}{!}{
\begin{tabular}{l|ccccccc}
\Xhline{4\arrayrulewidth}
\multicolumn{1}{c|}{\multirow{2}{*}{Method}} & Imagenet   & \multicolumn{6}{c}{MS-COCO}               \\
\multicolumn{1}{c|}{}                        & Top-1 Acc. & TR@1 & TR@5 & TR@10 & IR@1 & IR@5 & IR@10 \\ \hline
OneR                                         & 23.7       & 25.5 & 48.2 & 60.2  & 16.9 & 36.9 & 47.9  \\
OneR - MIM                                   & 23.4       & 24.7 & 46.9 & 58.7  & 16.9 & 36.8 & 47.7  \\
OneR - MIM - MLM                             & 22.9       & 23.3 & 47.2 & 58.1  & 13.6 & 33.3 & 45.6  \\ \hline
\end{tabular}}
\caption{Additional ablations on masked modeling objectives. All models are trained with CC3M.}
\end{table}


We present additional ablations of our framework. We observe that training with masked image modeling (MIM) and masked language modeling (MLM) helps the performance, but is not the most crucial component. 




\section{C. Implementation Details}
\:\\
We adopt the model architecture of Masked AutoEncoder~\cite{he2022masked} with BERT~\cite{devlin2018bert} word embeddings and language modeling head. Unlike most prior works on VLP, we initialize our entire model \textit{from scratch}, as neither ViT nor BERT suits our goal towards a unified VL representation space. 1D and 2D sinusoidal positional embeddings are added to text and image respectively, and a single \texttt{[CLS]} token is prepended to all three input types. 
Special modality indicator tokens (\textit{e.g.,} \texttt{[SEP]} or \texttt{[SEG]}) are further removed from typical one tower baselines in order to train a fully modality-agnostic representation learner. 

\begin{figure}[t]
\centering
  \includegraphics[width=0.98\linewidth]{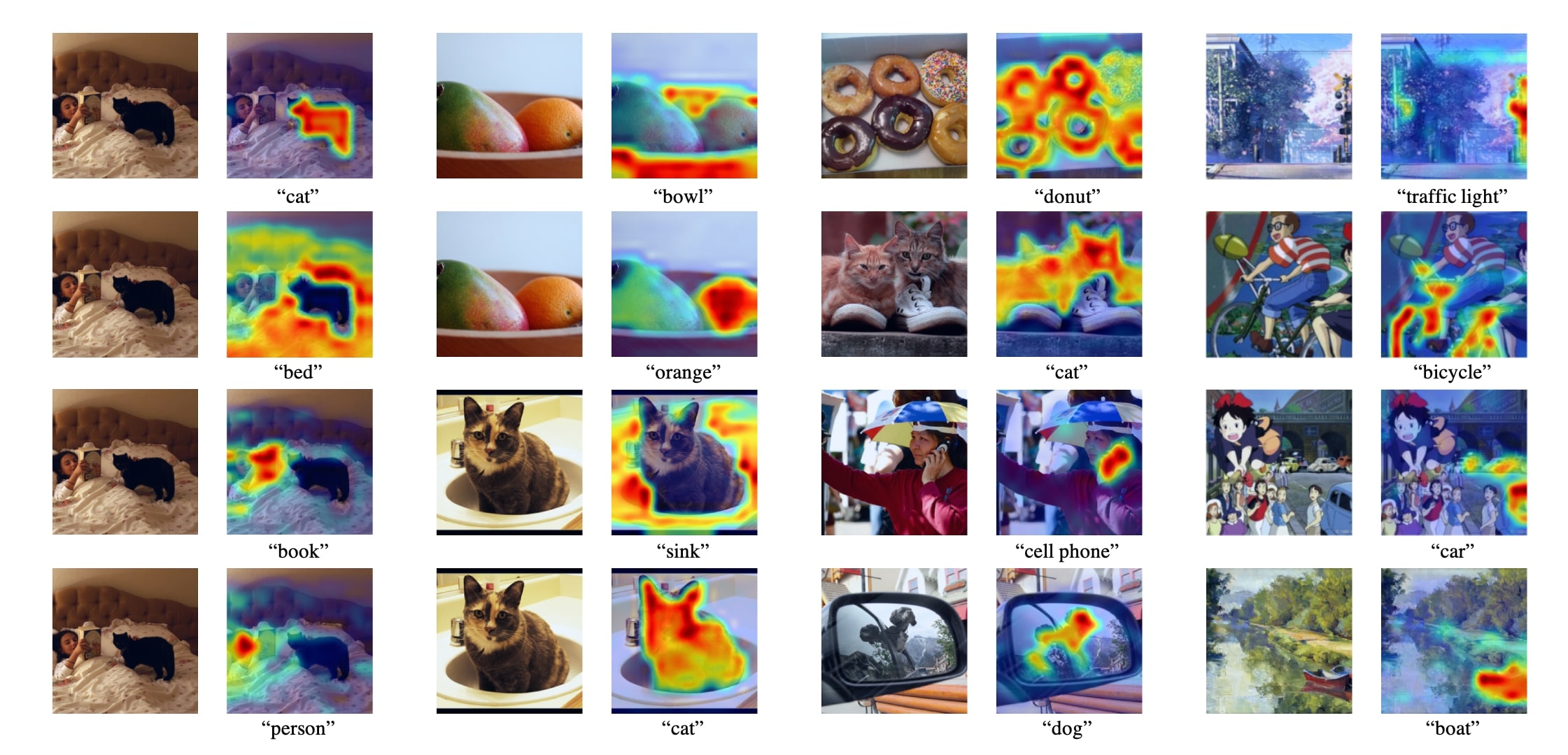}
  \caption{Additional zero-shot localization results. We compute cosine similarity between image patches and the text prompt, highlighting the related regions in a simple and straightforward manner.}
  \label{fig:supp_qual}
\end{figure}

\begin{figure}[t]
\centering
  \includegraphics[width=0.98\linewidth]{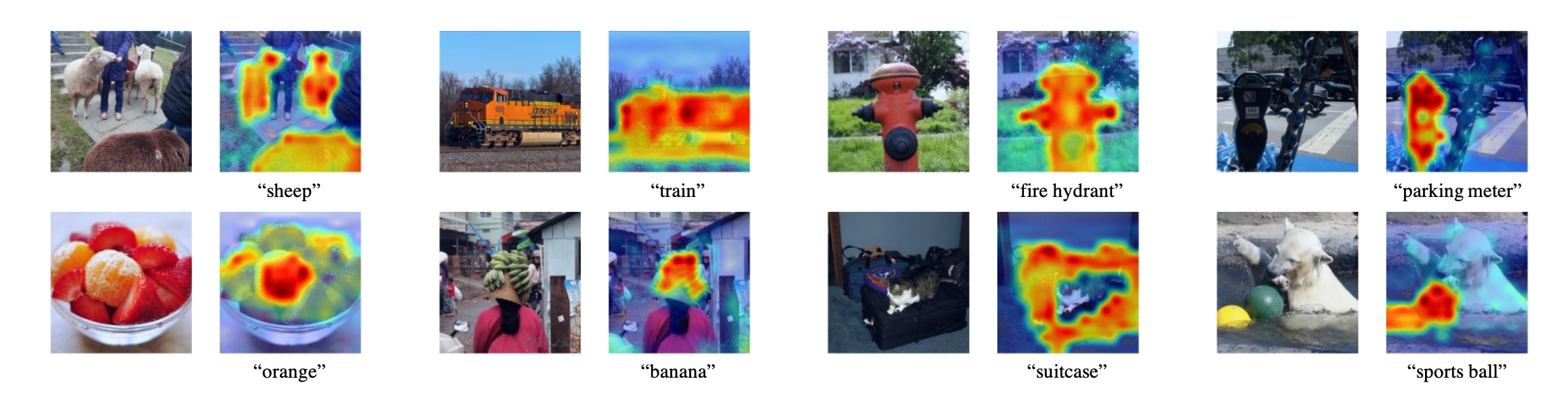}
  \caption{Additional zero-shot localization results by the definition. We compute cosine similarity between image patches and the text prompt (definition), highlighting the related regions in a simple and straightforward manner.}
  \label{fig:supp_qual2}
\end{figure}

Overall, we follow the settings of MoCo-v3~\cite{chen2021empirical}, but with learnable ConvStem~\cite{xiao2021early} as the image patch projector. A 3-layer MLP projector and a 2-layer predictor are used as in \cite{chen2020simple}, and momentum ratio was fixed to 0.996 throughout the whole training process. We choose the base learning rate of 1e-4 with linear scaling rule \cite{krizhevsky2014one, goyal2017accurate} that adapts the learning rate as lr$\times$BatchSize/256. First 4 epochs are for warming up and cosine scheduling~\cite{loshchilov2016sgdr} decays the lr for a total of 40 epochs. After 40 epochs of training on 224$\times$224 resolution images, we further train with 384$\times$384 upsampled resolution for additional 5 epochs with positional embeddings interpolated correspondingly. Batch size is 4,096 for 224$\times$224 stage and 1,024 for 384$\times$384. We optimize with AdamW~\cite{loshchilov2017decoupled} under the weight decay of 0.1.


Unlike recent works~\cite{li2021align, yang2022vision} that go through additional forward passes for masked modeling during training, OneR computes the contrastive loss and the masked modeling loss simultaneously in a single forward pass. Only the online network learns masked modeling, thus clean inputs are fed to the momentum model. MLM masking ratio is set to $0.15$ as done in \cite{li2021align, devlin2018bert}, and MIM ratio is raised from $0.1$ to $0.5$.


\section{D. Qualitative Results}

We present additional patch embedding similarity visualizations. \cref{fig:supp_qual} displays results from using the object category as the text query, and \cref{fig:supp_qual2} shows similarity map by the object definition.

\section{E. Discussions}

\textbf{Random Augmentation}
In our early experiments, we have observed that adding random augmentations does not help the overall model performance, so we train OneR without strong augmentations. We conjecture this could be due to the difficulty of the task: learning a single vision-language representation space. As we have discussed in the main paper, our basic hypothesis is that a paired image and text can be seen as two different (but closely related) views of an underlying semantic. Hence, additionally performing strong augmentations on one side could be unnecessary. Nevertheless, we believe that there are much room for more sophisticated designs to incorporate data augmentation into the framework, which we leave to future works.

\noindent
\textbf{Momentum Teacher}
As we remove strong image augmentations, the presence of a momentum teacher is critical to the performance of OneR. Although it is a relatively common belief that contrastive learning with a momentum teacher network improves the performance~\cite{li2021align, yang2022vision}, we observe that it matters more in a unified single-tower setting. Exploration into such behaviors could be another promising research direction.


\end{document}